%% file: main.tex
\pdfoutput=1
\documentclass[letterpaper, 10 pt, conference]{ieeeconf}  

\IEEEoverridecommandlockouts                              

\usepackage{algorithm} 
\usepackage{algpseudocode} 
\usepackage{hyperref}
\usepackage{amssymb}
\usepackage{amsmath}
\usepackage{bbm}
\usepackage{url}
\usepackage{graphicx}
\usepackage{subfigure}
\usepackage{wrapfig}
\usepackage{makecell}
\usepackage{color,array}
\usepackage{bm}

\title{\LARGE \bf
Option-Aware Adversarial Inverse Reinforcement Learning \\for Robotic Control
}

\author{Jiayu Chen$^{1}$, Tian Lan$^{2}$, and Vaneet Aggarwal$^{1}$
\thanks{$^{1}$J. Chen and V. Aggarwal are with the School of Industrial Engineering,
        Purdue University, West Lafayette, IN 47907, USA
        {\tt\small chen3686@purdue.edu, vaneet@purdue.edu}. V. Aggarwal is also with the CS Department, KAUST, Thuwal, Saudi Arabia. }%
        \thanks{$^{2}$T. Lan is with the Department of Electrical and Computer Engineering, George Washington University, Washington D.C., 20052, USA
        {\tt\small tlan@gwu.edu}}%
}

\begin{document}

\maketitle
\thispagestyle{empty}
\pagestyle{empty}

\input{abstract}

\input{introduction}

\input{related_work}

\input{background}

\input{algorithm}

\input{evaluation}

\input{conclusion}

\clearpage
\bibliographystyle{./IEEEtran} 
\bibliography{./IEEEabrv,./IEEEexample}

\clearpage
\input{appendix}

\end{document}

%% file: abstract.tex
\begin{abstract}
Hierarchical Imitation Learning (HIL) has been proposed to recover highly-complex behaviors in long-horizon tasks from expert demonstrations by modeling the task hierarchy with the option framework. Existing methods either overlook the causal relationship between the subtask and its corresponding policy or cannot learn the policy in an end-to-end fashion, which leads to suboptimality. In this work, we develop a novel HIL algorithm based on Adversarial Inverse Reinforcement Learning and adapt it with the Expectation-Maximization algorithm in order to directly recover a hierarchical policy from the unannotated demonstrations. Further, we  introduce a directed information term to the objective function to enhance the causality and propose a Variational Autoencoder framework for learning with our objectives in an end-to-end fashion. Theoretical justifications and evaluations on challenging robotic control tasks are provided to show the superiority of our algorithm. The codes are available at \href{https://github.com/LucasCJYSDL/HierAIRL}{https://github.com/LucasCJYSDL/HierAIRL}.
\end{abstract}

%% file: introduction.tex
\section{INTRODUCTION}

Reinforcement Learning (RL) has achieved impressive performance in a variety of scenarios, such as games \cite{brown2019superhuman, DBLP:journals/nature/SilverHMGSDSAPL16,DBLP:journals/corr/HosuR16} and robotic control \cite{ebert2018visual, lillicrap2015continuous}. However, most of its applications rely on carefully-crafted, task-specific reward signals to drive exploration and learning, limiting its use in real-life scenarios. In this case, Imitation Learning (IL) methods have been developed to acquire a policy for a certain task based on the corresponding expert demonstrations (e.g., trajectories of state-action pairs) rather than reinforcement signals. However, complex long-horizon tasks can often be broken down and processed as a series of subtasks which can serve as basic skills for completing various compound tasks. In this case, learning a single monolithic policy with IL to represent a structured activity can be challenging. Therefore, Hierarchical Imitation Learning (HIL) has been proposed to recover a two-level policy for a long-horizon task from the demonstrations. Specifically, HIL trains low-level policies (i.e., skills) for accomplishing the specific control for each sub-task, and a high-level policy for scheduling the switching of the skills. Such a hierarchical policy, which is usually formulated with the option framework \cite{DBLP:journals/ai/SuttonPS99}, makes full use of the sub-structure between the parts within the activity and has the potential for better performance.

The most state-of-the-art (SOTA) works on HIL \cite{DBLP:conf/iclr/SharmaSRK19,DBLP:conf/icml/JingH0MKGL21} are developed based on Generative Adversarial Imitation Learning (GAIL) \cite{ho2016generative} which is a widely-adopted IL algorithm. In \cite{DBLP:conf/iclr/SharmaSRK19}, they additionally introduce a directed information \cite{massey1990causality} term to the GAIL objective function. In this way, their method can enhance the causal relationship between the skill choice and the corresponding state-action sequence to form low-level policies for each subtask, while encouraging the hierarchical policy to generate trajectories similar to the expert ones in distribution through the GAIL objectives. However, they update the high-level and low-level policy on two separate stages. Specifically, the high-level policy is learned with behavioral cloning, which is a supervised IL algorithm and vulnerable to compounding errors \cite{ross2011reduction}, and remains fixed during the low-level policy learning with GAIL. Given that the two-level policies are coupled with each other, such a two-staged paradigm potentially leads to sub-optimal solutions. On the other hand, in \cite{DBLP:conf/icml/JingH0MKGL21}, they propose to learn a hierarchical policy by option-occupancy measurement matching, that is, imitating the joint distribution of the options, states and actions of the expert demonstrations rather than only matching the state-action distributions like GAIL. However, they overlook the causal relationship between the subtask structure and the policy hierarchy, so the recovered policy may degenerate into a poorly-performed monolithic policy for the whole task, especially when the option annotations of the expert demonstrations are not provided. 
 
 In this work, we propose a novel HIL algorithm -- Hierarchical Adversarial Inverse Reinforcement Learning (H-AIRL), which integrates the SOTA IL algorithm Adversarial Inverse Reinforcement Learning (AIRL) \cite{DBLP:journals/corr/abs-1710-11248} with the option framework through an objective based on the directed information. 
 Compared with GAIL, AIRL is able to recover the expert reward function along with the expert policy and has more robust and stable performance for challenging robotic tasks \cite{DBLP:journals/corr/abs-1710-11248, jena2021augmenting, wang2021decision}. From the algorithm perspective, our contributions are as follows: \textbf{(1)} We propose a practical lower bound of the directed information between the option choice and the corresponding state-action sequences, which can be modeled as a variational posterior and updated in a Variational Autoencoder (VAE) \cite{DBLP:journals/corr/KingmaW13} framework. This design enables our algorithm to update the high-level and low-level policy at the same time in an end-to-end fashion, which is an improvement compared with \cite{DBLP:conf/iclr/SharmaSRK19}. \textbf{(2)} We redefine the AIRL objectives on the extended state and action space, in order to directly recover a hierarchical policy from the demonstrations. \textbf{(3)} We provide an Expectation-Maximization (EM) \cite{moon1996expectation} adaption of our algorithm so that it can be applied to the expert demonstrations without the sub-task annotations (i.e., unsegmented expert data) which are easier to obtain in practice. Further, we provide solid theoretical justification of the three folds mentioned above, and comparisons of our algorithm with SOTA HIL and IL baselines on multiple Mujoco \cite{todorov2012mujoco} continuous control tasks where our algorithm significantly outperforms the others.

%% file: related_work.tex
\section{RELATED WORK}

\textbf{Imitation Learning.} Imitation learning methods \cite{DBLP:journals/ras/ArgallCVB09} seek to learn to perform a task from expert demonstrations, where the learner is given only samples of trajectories from the expert and is not provided any reinforcement signals, such as the environmental rewards which are usually hard to acquire in real-life scenarios. There are two main branches for this algorithm setting: behavioral cloning (BC) \cite{DBLP:journals/neco/Pomerleau91}, which learns a policy as a supervised learning problem over state-action pairs from expert trajectories, and inverse reinforcement learning (IRL) \cite{DBLP:conf/icml/NgR00}, which first infers a reward function under which the expert is uniquely optimal and then recovers the expert policy based on it. Behavioral cloning only tends to succeed with large amounts of data, due to the compounding error caused by covariate shift \cite{DBLP:journals/jmlr/RossGB11}. Inverse reinforcement learning, while avoiding the compounding error, is extremely expensive to solve and scale, since it requires reinforcement learning to get the corresponding optimal policy in each iteration of updating the reward function. GAIL \cite{ho2016generative} and AIRL \cite{DBLP:journals/corr/abs-1710-11248} have been proposed to scale IRL for complex high-dimensional control tasks. They realize IRL through an adversarial learning framework, where they alternatively update a policy and discriminator network. The discriminator serves as the reward function and learns to differentiate between the expert demonstrations and state-action pairs from the learned policy. While, the policy is trained to generate trajectories that are difficult to be distinguished from expert data by the discriminator. Mathematical details are provided in Section \ref{airl}. AIRL explicitly recovers the reward function and provides more robust and stable performance among challenging tasks \cite{DBLP:journals/corr/abs-1710-11248, jena2021augmenting, wang2021decision}, which is chosen as our base algorithm for extension.

\textbf{Hierarchical Imitation Learning.} Given the nature of subtask decomposition in long-horizon tasks, hierarchical imitation learning can achieve better performance than imitation learning by forming micro-policies for accomplishing the specific control for each subtask first and then learning a macro-policy for scheduling among the micro-policies. The micro-policies (a.k.a., skills) in RL can be modeled with the option framework proposed in \cite{DBLP:journals/ai/SuttonPS99}, which extends the usual notion of actions to include options — the closed-loop policies for taking actions over a period of time. We provide further details about the option framework in Section \ref{option}. Through integrating IL with the options, the hierarchical versions of the IL methods mentioned above have been developed, including hierarchical behavioral cloning (HBC) and  hierarchical inverse reinforcement learning (HIRL). In HBC, they train a policy for each subtask through supervised learning with the corresponding state-action pairs, due to which the subtask annotations need to be provided or inferred. In particular, the methods proposed in \cite{DBLP:conf/icml/0001JADYD18, DBLP:conf/icml/KipfLDZSGKB19} require segmented data with the subtask information. While, in \cite{DBLP:journals/ml/DanielHPN16, DBLP:conf/aistats/ZhangP21}, they infer the subtask information as the hidden variables in a Hidden Markov Model \cite{fine1998hierarchical} and solve the HBC as an MLE problem with the Expectation–Maximization (EM) algorithm \cite{moon1996expectation}. Despite its theoretical completeness, HBC is also vulnerable to compounding errors in case of limited demonstrations. On the other hand, the HIRL methods proposed in \cite{DBLP:conf/iclr/SharmaSRK19, DBLP:conf/icml/JingH0MKGL21} have extended GAIL with the option framework to recover the hierarchical policy (i.e., the high-level and low-level policies mentioned above) from unsegmented expert data. Specifically, in \cite{DBLP:conf/iclr/SharmaSRK19}, they introduce a regularizer into the original GAIL objective function to maximize the directed information between generated trajectories and the subtask/option annotations. However, the high-level and low-level policies are trained in two separate stages in their approach, which will inevitably lead to convergence with a poor local optimum. As for the approach proposed in \cite{DBLP:conf/icml/JingH0MKGL21} which claims to outperform \cite{DBLP:conf/iclr/SharmaSRK19} and HBC, it replaces the occupancy measurement in GAIL, which measures the distribution of the state-action pairs, with option-occupancy measurement to encourage the hierarchical policy to generate state-action-option tuples with similar distribution to the expert demonstrations. However, they do not adopt the directed information objective to enhance the causal relationship between the option choice and the corresponding state-action sequence. In this paper, we propose a new HIL algorithm based on AIRL, which takes advantage of the directed information objective and updates the high-level and low-level policies in an end-to-end fashion. Moreover, we provide  theoretical justification of our algorithm, and demonstrate its superiority on challenging robotic control tasks.

%% file: background.tex
\section{BACKGROUND}

In this section, we introduce the background knowledge of our work, including AIRL and the One-step Option Framework. These are defined with the Markov Decision Process (MDP), denoted by $\mathcal{M}=(\mathcal{S}, \mathcal{A}, \mathcal{P}, \mu, \mathcal{R}, \mathcal{\gamma})$, where $\mathcal{S}$ is the state space, $\mathcal{A}$ is the action space, $\mathcal{P}:\mathcal{S} \times \mathcal{A} \times \mathcal{S} \rightarrow [0,1]$ is the state transition function, $\mu:\mathcal{S}\rightarrow [0,1]$ is the distribution of the initial state, $\mathcal{R}:\mathcal{S} \times \mathcal{A} \rightarrow \mathbb{R}$ is the reward function, and $\mathcal{\gamma} \in (0,1]$ is the discount factor. 

\subsection{Adversarial Inverse Reinforcement Learning} \label{airl}

Inverse reinforcement learning (IRL) \cite{DBLP:conf/icml/NgR00} aims to infer an expert's reward function from demonstrations, based on which the policy of the expert can be recovered. As a representative, Maximum Entropy IRL \cite{DBLP:conf/aaai/ZiebartMBD08} solves it as a maximum likelihood estimation (MLE) problem shown as Equation (\ref{equ:1}). $\tau_{E} \triangleq (S_0, A_0, \cdots, S_{T-1}, A_{T-1}, S_T)$ denotes the expert trajectory, i.e., a sequence of state-action pairs of horizon $T$. $Z_{\vartheta}$ is the partition function defined as $Z_{\vartheta}=\int\widehat{P}_{\vartheta}(\tau_{E})d\tau_{E}$ (continuous $\mathcal{S}$ and $\mathcal{A}$) or $Z_{\vartheta}=\sum_{\tau_{E}}\widehat{P}_{\vartheta}(\tau_{E})$ (discrete $\mathcal{S}$ and $\mathcal{A}$).
\begin{equation} \label{equ:1}
\begin{aligned}
&\qquad\mathop{max}_{\vartheta}\mathbb{E}_{\tau_{E}}\left[\log P_{\vartheta}(\tau_{E})\right],\ P_{\vartheta}(\tau_{E})=\widehat{P}_{\vartheta}(\tau_{E})/Z_{\vartheta} \\
    &\widehat{P}_{\vartheta}(\tau_{E})=\mu(S_0)\mathop{\prod}_{t=0}^{T-1}\mathcal{P}(S_{t+1}|S_{t}, A_{t})\exp(\mathcal{R}_{\vartheta}(S_t, A_t))
\end{aligned}
\end{equation}

Since $Z_{\vartheta}$ is intractable for the large-scale state-action space, the authors of \cite{DBLP:journals/corr/abs-1710-11248} propose Adversarial Inverse Reinforcement Learning (AIRL) to solve this MLE problem in a sample-based manner. They realize this through alternatively training a discriminator $D_{\vartheta}$ and policy network $\pi$ in an adversarial setting. Specifically, the discriminator is trained by minimizing the cross-entropy loss between the expert demonstrations $\tau_{E}$ and generated samples $\tau$ by $\pi$:
\begin{equation} \label{equ:3}
    \mathop{min}_{\vartheta}-\sum_{t=0}^{T-1}\mathbb{E}_{\tau_{E}}[\log D_{\vartheta}(S_t,A_t)] - \mathbb{E}_{\tau}[\log (1-D_{\vartheta}(S_t,A_t))]
\end{equation}
where $D_{\vartheta}(S,A)=\exp(f_{\vartheta}(S,A))/[\exp(f_{\vartheta}(S,A))+\pi(A|S)]$. Meanwhile, the policy $\pi$ is trained with off-the-shelf RL algorithms using the reward function defined as $\log D_{\vartheta}(S,A)-\log(1-D_{\vartheta}(S,A))$. Further, they justify that, at optimality, $f_{\vartheta}(S,A)$ can serve as the recovered reward function $\mathcal{R}_{\vartheta}(S,A)$ and $\pi$ is the recovered expert policy which maximizes the entropy-regularized objective: $\mathbb{E}_{\tau \sim \pi}\left[\mathop{\sum}_{t=0}^{T-1}\mathcal{R}_{\vartheta}(S_t,A_t)-\log\pi(A_t|S_t)\right]$.

\subsection{One-step Option Framework} \label{option}


As proposed in \cite{DBLP:journals/ai/SuttonPS99}, an option $Z \in \mathcal{Z}$ can be described with three components: an initiation set $I_Z \subseteq \mathcal{S}$, an intra-option policy $\pi_Z(A|S): \mathcal{S} \times \mathcal{A} \rightarrow [0,1]$, and a termination function $ \beta_Z(S): \mathcal{S} \rightarrow [0,1]$. An option $Z$ is available in state $S$ if and only if $S \in I_Z$. Once the option is taken, actions are selected according to $\pi_Z$ until it terminates stochastically according to $\beta_Z$, i.e., the termination probability at the current state. A new option will be activated in this call-and-return
style by a high-level policy $\pi_\mathcal{Z}(Z|S): \mathcal{S} \times \mathcal{Z} \rightarrow [0,1]$ once the last option terminates. In this way, $\pi_\mathcal{Z}(Z|S)$ and $\pi_Z(A|S)$ constitute a hierarchical policy for a certain task. However, it's inconvenient to deal with the initiation set $I_Z$ and termination function $\beta_Z$ while learning this hierarchical policy. Thus, in \cite{DBLP:conf/nips/ZhangW19a, DBLP:conf/icml/JingH0MKGL21}, they adopt the one-step option framework. It's assumed that each option is available in each state, i.e., $I_Z=\mathcal{S}, \forall Z \in \mathcal{Z}$. Also, the high-level and low-level (i.e., intra-option) policy are redefined as $\pi_{\theta}$ and $\pi_{\phi}$ respectively:
\begin{equation} \label{equ:4}
\begin{aligned}
    &\pi_{\theta}(Z|S, Z')=\beta_{Z'}(S)\pi_{\mathcal{Z}}(Z|S) + \left[(1-\beta_{Z'}(S))\delta_{Z=Z'}\right] \\ 
    &\qquad\qquad\qquad\pi_{\phi}(A|S,Z)=\pi_Z(A|S)
\end{aligned}
\end{equation}
where $Z'$ denotes the option in the last time step and $\delta_{Z=Z'}$ is the indicator function. We can see that if the previous option terminates (with probability $\beta_{Z'}(S)$), the agent will select a new option according to $\pi_{\mathcal{Z}}(Z|S)$; otherwise, it will stick to $Z'$. With the new definition and assumption, we can optimize the hierarchical policy $\pi_{\theta}$ and $\pi_{\phi}$ without the extra need to justify the exact beginning and breaking condition of each option. Nevertheless, $\pi_{\theta}(Z|S, Z')$ still includes two separate parts, i.e., $\beta_{Z'}(S)$ and $\pi_{\mathcal{Z}}(Z|S)$, and due to the indicator function, the update gradients of $\pi_{\mathcal{Z}}$ will be blocked/gated by the termination function $\beta_{Z'}(S)$. In this case, the authors of \cite{li2020skill} propose to marginalize the termination function away, and instead implement $\pi_{\theta}(Z|S, Z')$ as an end-to-end neural network (NN) with the Multi-Head Attention (MHA) mechanism \cite{DBLP:conf/nips/VaswaniSPUJGKP17} which enables their algorithm to temporally extend options in the absence of the termination function. We provide more details on MHA and the structure design of $\pi_{\theta}$ and $\pi_{\phi}$ in Appendix \ref{mha} \footnote{All the appendices are available in the extended version of our paper at \href{https://github.com/LucasCJYSDL/HierAIRL/blob/main/ICRA.pdf}{https://github.com/LucasCJYSDL/HierAIRL/blob/main/ICRA.pdf}}. With the marginalized one-step option framework, we only need to train the two NN-based policy, i.e., $\pi_{\theta}$ and $\pi_{\phi}$. In particular, we adopt the Hierarchical Reinforcement Learning algorithm, i.e., SA, proposed in \cite{li2020skill} to learn $\pi_{\theta}$ and $\pi_{\phi}$.

%% file: algorithm.tex
\section{PROPOSED APPROACH}

\subsection{Optimization with the Directed Information Objective}

\begin{figure}
\centering
\includegraphics[width=3.0in, height=1.1in]{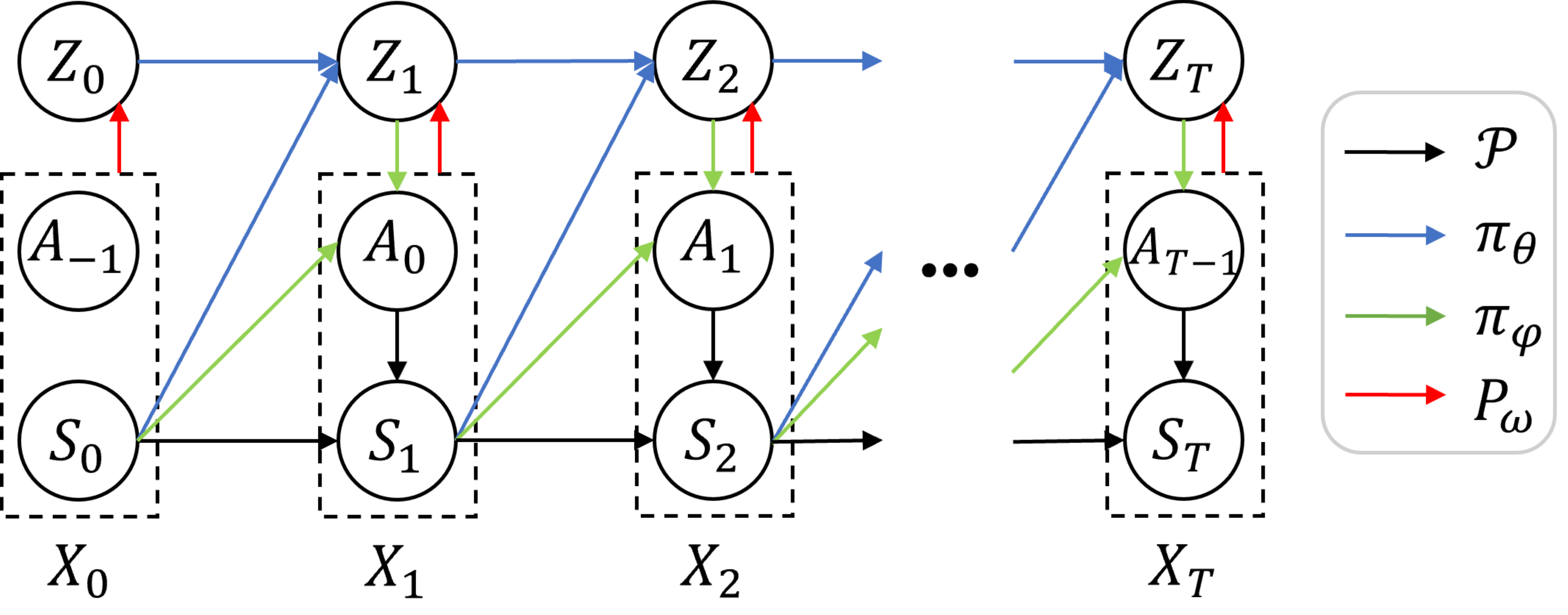}
\caption{Illustration of the probabilistic graphical model and its implementation with the one-step option model.}
\label{fig:1}
\end{figure}

Our work focuses on learning a hierarchical policy from expert demonstrations through integrating the one-step option framework with AIRL. In this section, we define the directed information objective function for training the hierarchical policy, fit it with the one-step option model, and propose how to optimize it in an end-to-end fashion with an RNN-based VAE structure, which is part of our novelty and contribution. 

As mentioned in Section \ref{option}, the hierarchical policy agent will first decide on its option choice $Z$ using the high-level policy $\pi_{\theta}$ and then select the primitive action based on the low-level policy $\pi_{\phi}$ corresponding to $Z$, when observing a new state. In this case, the policy learned should be conditioned on the option choice $Z$, and the option choice is specific to each timestep $t \in \{0,\cdots,T\}$, so we view the option choices $Z_{0:T}$ as the local latent contexts in a probabilistic graphical model shown as Figure \ref{fig:1}. It can be observed from Figure \ref{fig:1} that the local latent context $Z_{0:T}$ has a directed causal relationship with the trajectory $X_{0:T} = (X_0, \cdots, X_T) = ((A_{-1}, S_0), \cdots, (A_{T-1},S_T))$, where $A_{-1}$ is the dummy variable. Inspired by information theory \cite{massey1990causality, DBLP:conf/iclr/SharmaSRK19}, this kind of connection can be established by maximizing the directed information (a.k.a., casual information) flow from the trajectory to the latent factors of variation within the trajectory, i.e., $I(X_{0:T} \rightarrow Z_{0:T})$, which is defined as:
\begin{equation} \label{equ:13}
\begin{aligned}
        &I(X_{0:T} \rightarrow Z_{0:T}) = \mathop{\sum}_{t=1}^{T}\left[H(Z_t|Z^{t-1})-H(Z_t|X^{t},Z^{t-1})\right]\\
        &=\mathop{\sum}_{t=1}^{T}[H(Z_t|Z^{t-1}) +  \mathop{\sum}_{X^{t}, Z^{t}}P(X^{t}, Z^{t})\log P(Z_t|X^{t}, Z^{t-1})]
\end{aligned}
\end{equation}
where the first equality follows the general definition of directed information. Note that we use $X^t$ to represent $X_{0:t}$ for simplicity, and so on. In the above equations, $H(\cdot)$ denotes the entropy \cite{cover1999elements} and we assume the trajectory $X_{0:T}$ to be discrete in order to simplify the notations, but it can be either discrete or continuous in use. It is infeasible to directly optimize the above objective, since it is difficult to calculate the posterior distribution $P(Z_t|X_{0:t}, Z_{0:t-1})$ with only the hierarchical policy, i.e., $\pi_{\theta}$ and $\pi_{\phi}$. In this case, we instead maximize its variational lower bound as follows: (Please refer to Appendix \ref{dilb} for derivations.)
\begin{equation} \label{equ:15}
\begin{aligned}
        &L^{DI} \triangleq \mathop{\sum}_{t=1}^{T}[ H(Z_t|X^{t-1}, Z^{t-1}) + \\  &\mathop{\sum}_{\substack{X^t,Z^t}}P(X^t, Z^t)\log P_{\omega}(Z_t|X^t, Z^{t-1})]
\end{aligned}
\end{equation}
where $P_{\omega}$ is the variational estimation of $P(Z_t|X_{0:t}, Z_{0:t-1})$. While, $P(X_{0:t}, Z_{0:t})$ is calculated by: (Please refer to Appendix \ref{dilb} for derivations.)
\begin{equation} \label{equ:16}
\begin{aligned}
        &\mu(S_{0})\mathop{\prod}_{i=1}^{t}P(Z_{i}|X^{i-1}, Z^{i-1})P( A_{i-1}|X^{i-1}, Z^{i})\mathcal{P}_{S_{i-1},A_{i-1}}^{S_i}\\
        &=\mu(S_{0})\mathop{\prod}_{i=1}^{t}\pi_{\theta}(Z_{i}|S_{i-1}, Z_{i-1})\pi_{\phi}( A_{i-1}|S_{i-1}, Z_{i})\mathcal{P}_{S_{i-1},A_{i-1}}^{S_i}
\end{aligned}
\end{equation}
where $\mu$ is the initial state distribution and $\mathcal{P}_{S_{i-1},A_{i-1}}^{S_i}=\mathcal{P}(S_i|S_{i-1},A_{i-1})$ is the transition function. Note that the expectation term in $L^{DI}$ can be estimated through Monte Carlo sampling \cite{sutton2018reinforcement} by interacting with the environment using the definition in Equation \eqref{equ:16}, even though $\mu$ and $\mathcal{P}$ are unknown to us. Moreover, the second equality in Equation \eqref{equ:16} holds if we adopt the Markov assumption \cite{rabiner1986introduction}, i.e., the conditional probability distribution of a random variable depends only on its parent nodes in the probabilistic graphical model (e.g., Figure \ref{fig:1}). As a result, we can fit in the one-step option model (i.e., Equation \eqref{equ:4}) introduced in Section \ref{option}, and use $L^{DI}$ as the objective to update the hierarchical policy.

\begin{figure}
\centering
\includegraphics[width=3.0in, height=1.0in]{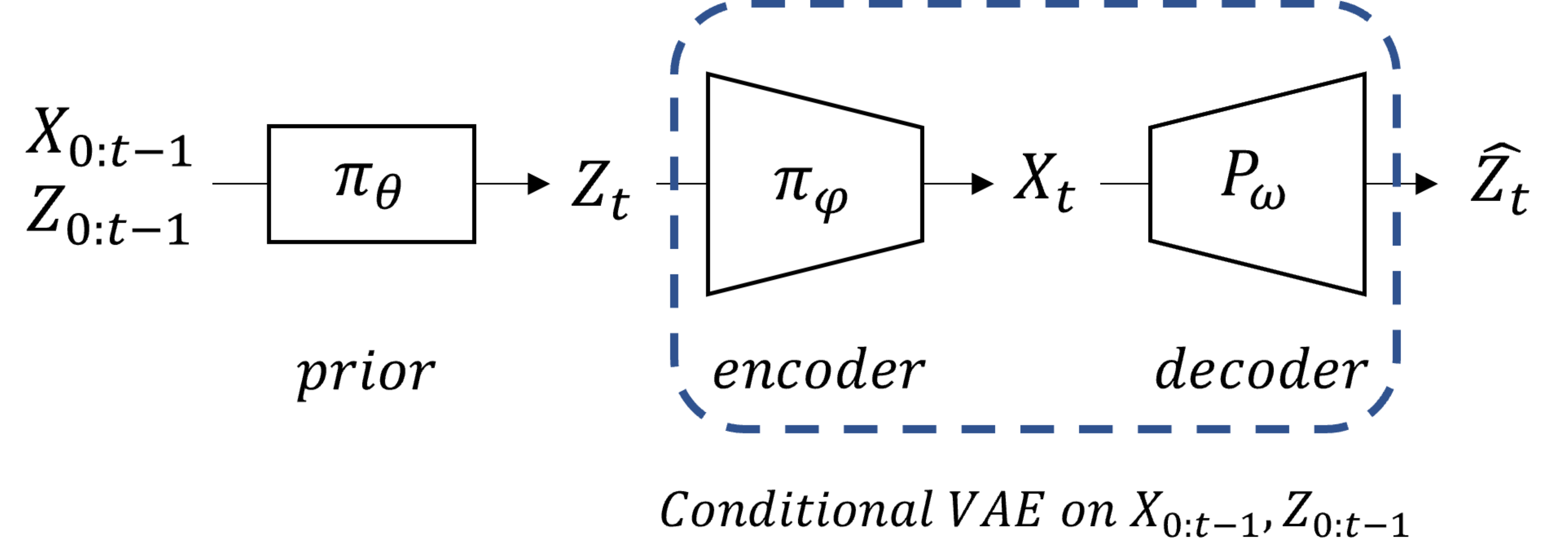}
\caption{The analogy of our learning framework with the VAE structure.}
\label{fig:2}
\end{figure}

To sum up, we can train the hierarchical policy $\pi_{\theta}$ and $\pi_{\phi}$ by maximizing the directed information to enhance the directed causal relationship between the latent contexts $Z_{0:T}$ and trajectories $X_{0:T}$ shown as Figure \ref{fig:1}. To realize this, we additionally introduce a variational posterior $P_{\omega}$ and update it together with $\pi_{\theta}$ and $\pi_{\phi}$ in a Variational Autoencoder (VAE) \cite{DBLP:journals/corr/KingmaW13} framework. As shown in Figure \ref{fig:2}, for the optimization of $L^{DI}$ (Equation \ref{equ:15}) which is a VAE-like objective function, at each timestep $t$, $\pi_{\phi}$ and $P_{\omega}$ form a conditional VAE between $Z_t$ and $X_t$, which is conditioned on the history, i.e., $(X_{0:t-1},Z_{0:t-1})$, with the prior distribution of $Z_t$ provided by $\pi_{\theta}$. Note that $P_{\omega}$ uses sequential data, i.e., $(X_{0:t},Z_{0:t-1})$, as input and thus are implemented with RNN, of which the details are in Appendix \ref{rnn-imp}. With this framework, we can train the hierarchical policy in an end-to-end manner, instead of learning the high- and low-level policy separately in two training schemes like in \cite{DBLP:conf/iclr/SharmaSRK19}. Moreover, in \cite{DBLP:conf/iclr/SharmaSRK19}, they overlook the first entropy term in the first line of Equation \eqref{equ:13} directly and get a negative lower bound of the positive directed information, which is trivial to solve.

\begin{figure*}[htbp]
\centering
\subfigure[Hopper]{
\label{fig:3(a)} 
\includegraphics[width=1.6in, height=1.3in]{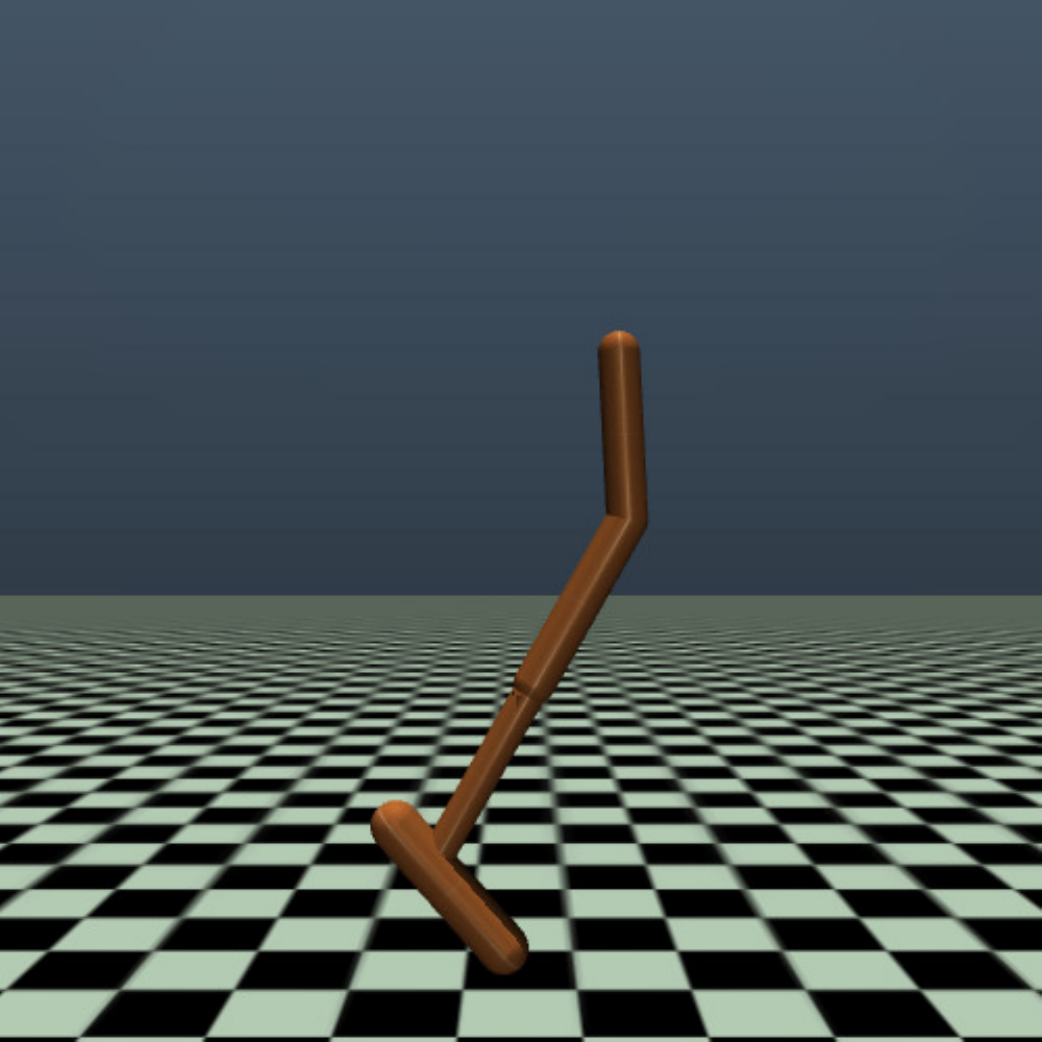}}
\subfigure[Walker]{
\label{fig:3(b)} 
\includegraphics[width=1.6in, height=1.3in]{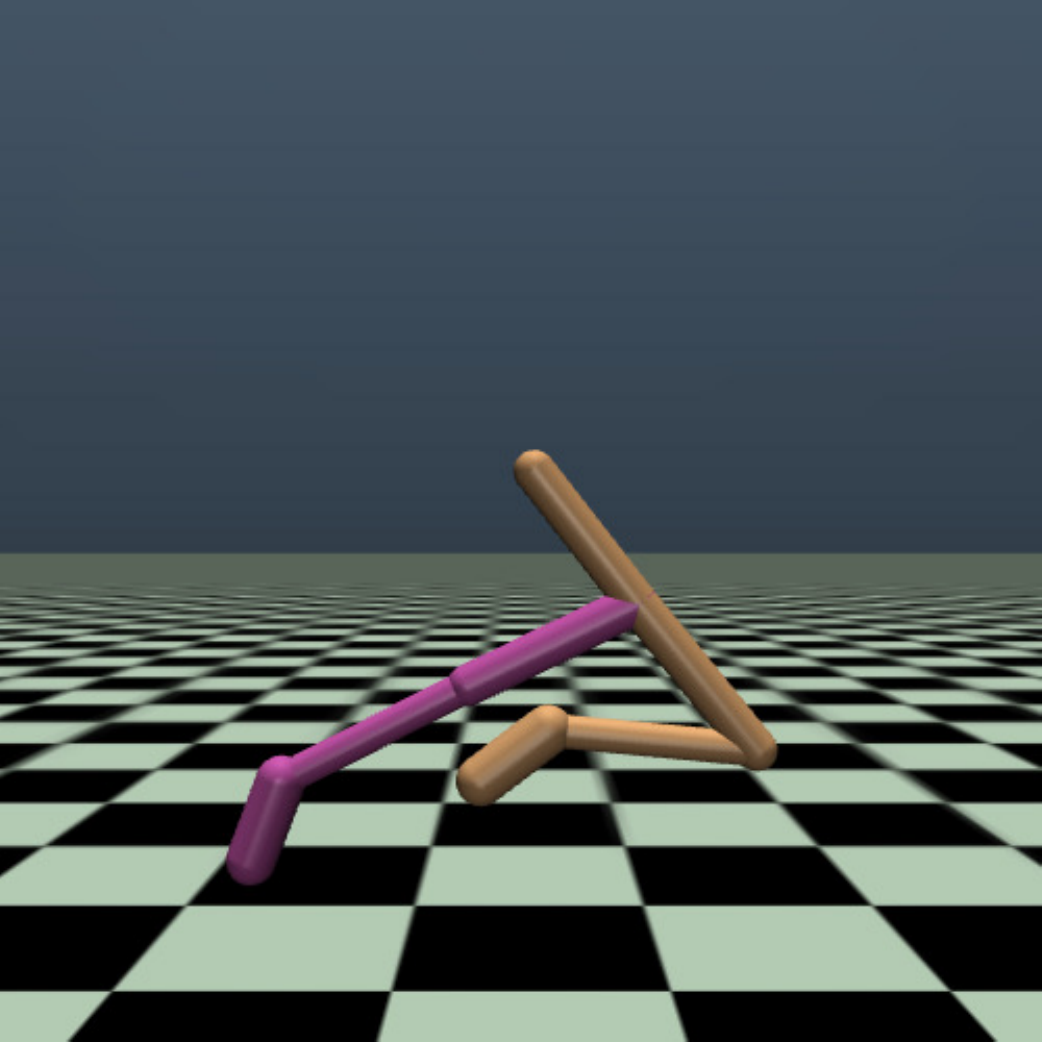}}
\subfigure[Ant]{
\label{fig:3(c)} 
\includegraphics[width=1.6in, height=1.3in]{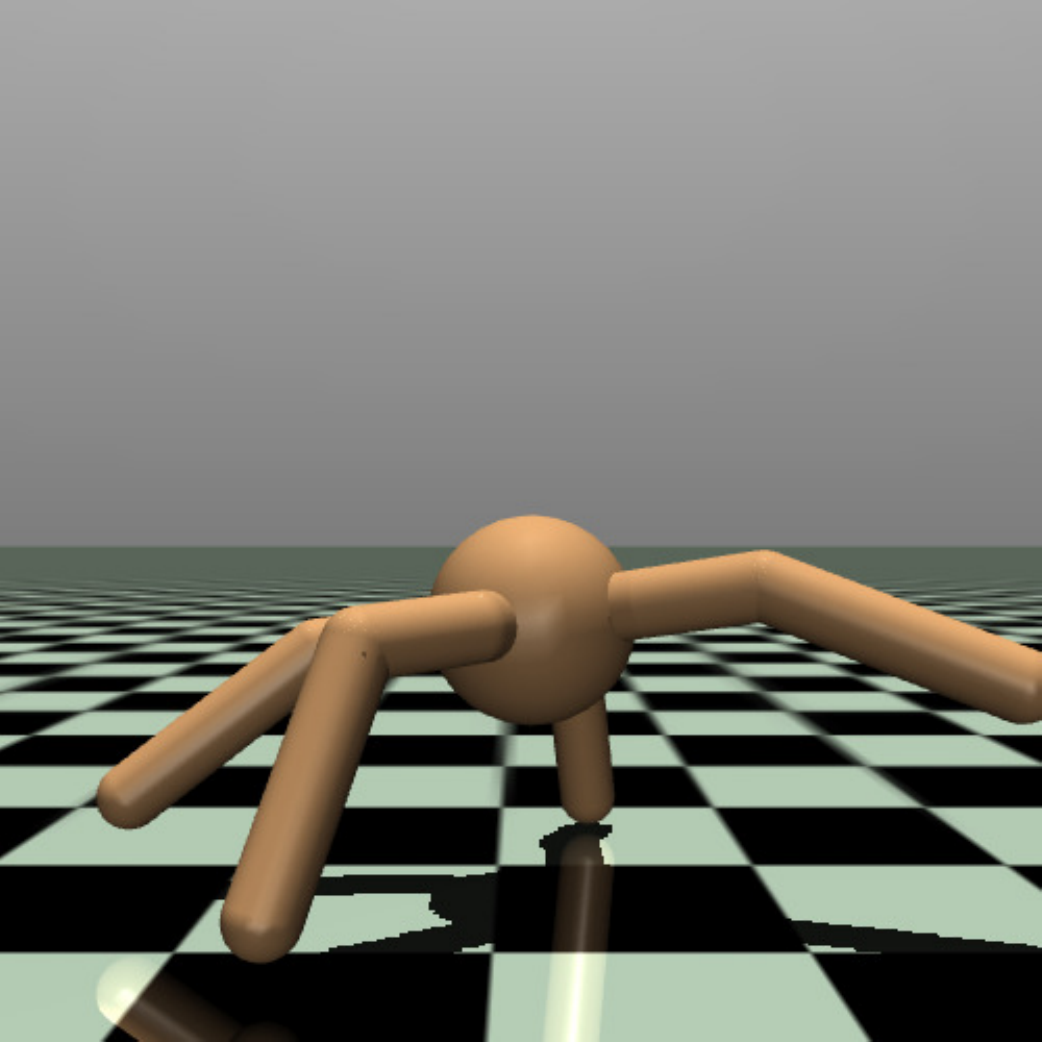}}
\subfigure[AntPusher]{
\label{fig:3(d)} 
\includegraphics[width=1.6in, height=1.3in]{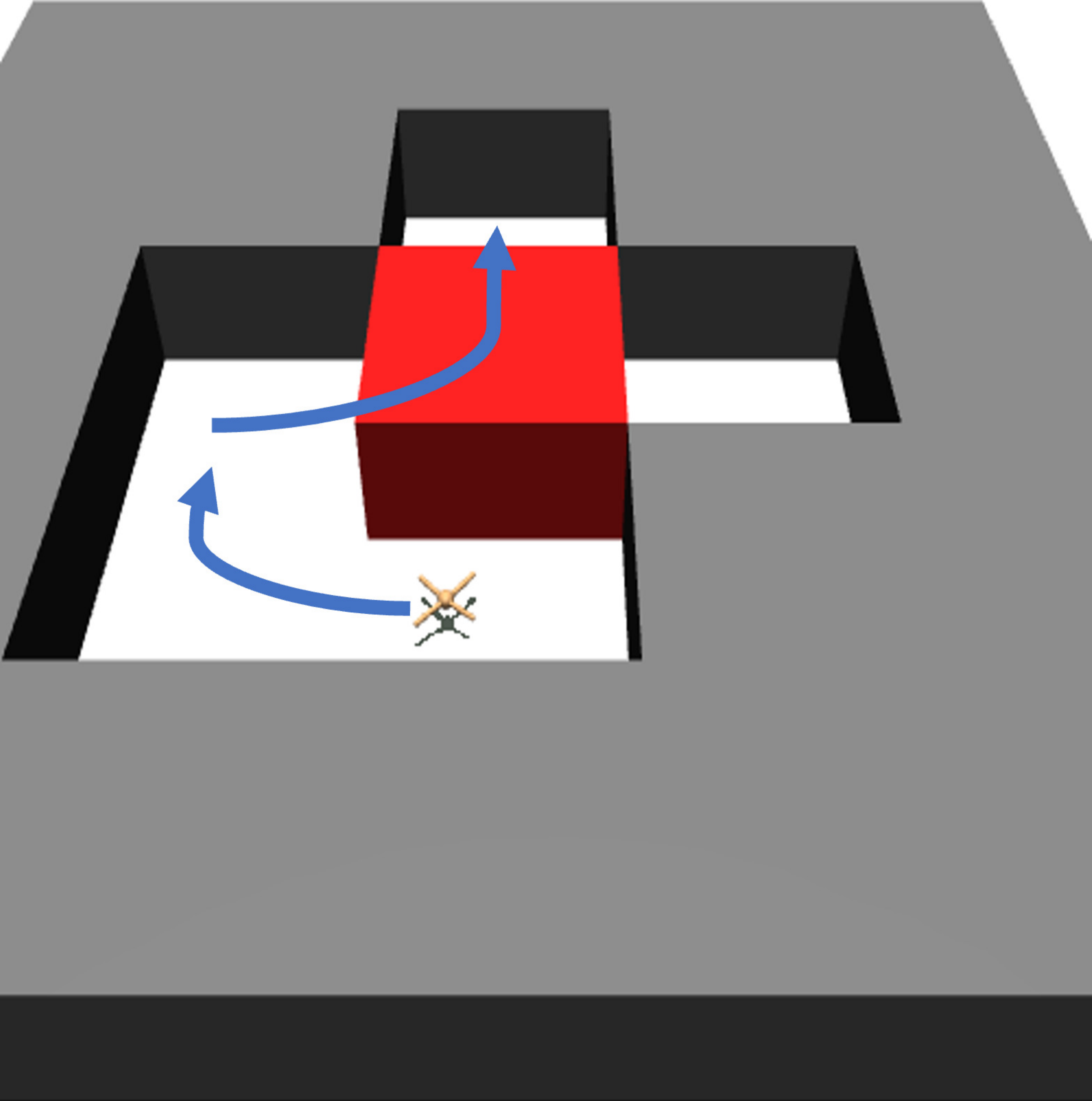}}
\caption{Evaluation tasks built with Mujoco.}
\label{fig:3} 
\end{figure*}

\begin{figure*}[htbp]
\centering
\subfigure[Hopper]{
\label{fig:4(a)} 
\includegraphics[width=2.25in, height=1.3in]{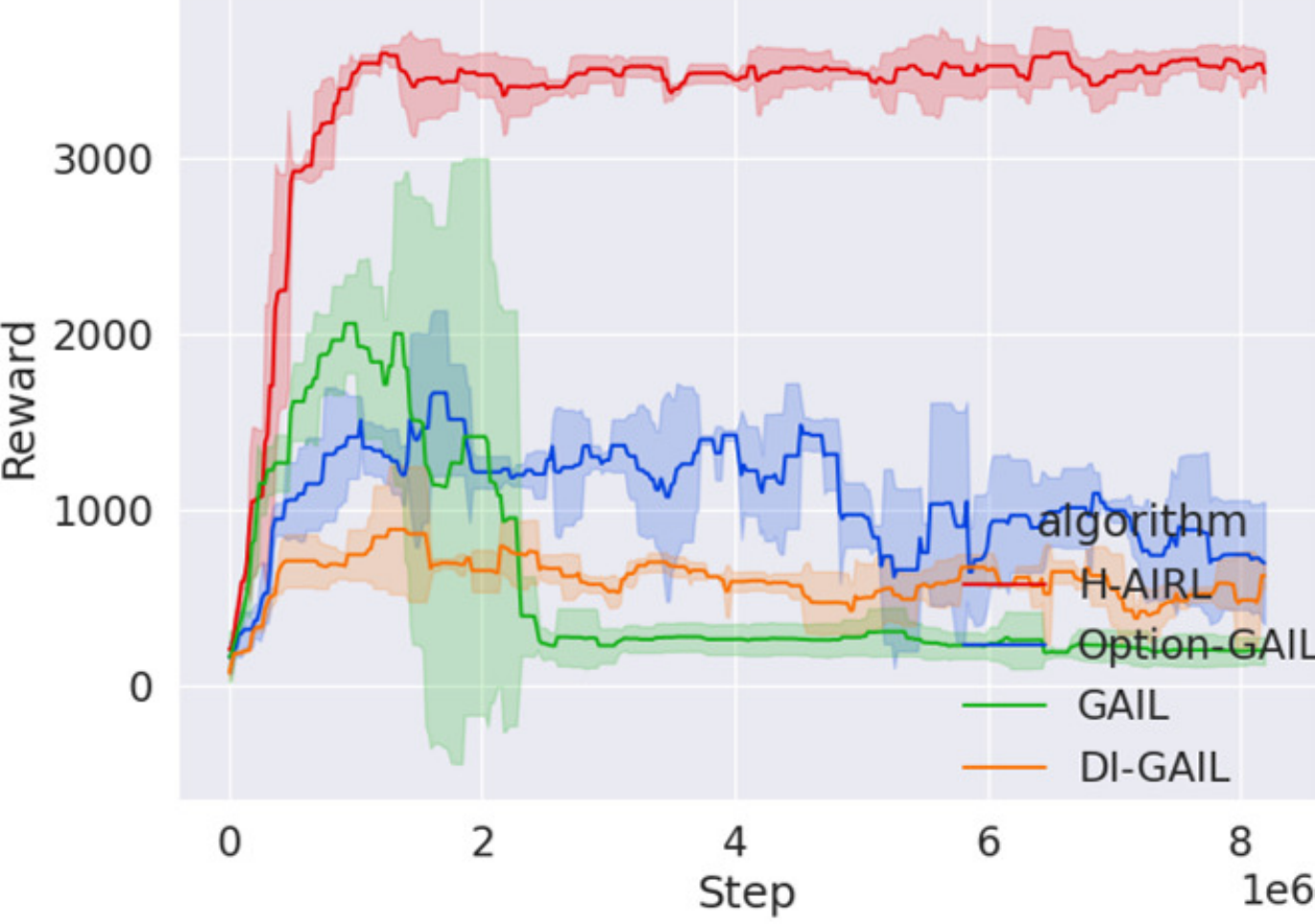}}
\subfigure[Walker]{
\label{fig:4(b)} 
\includegraphics[width=2.25in, height=1.3in]{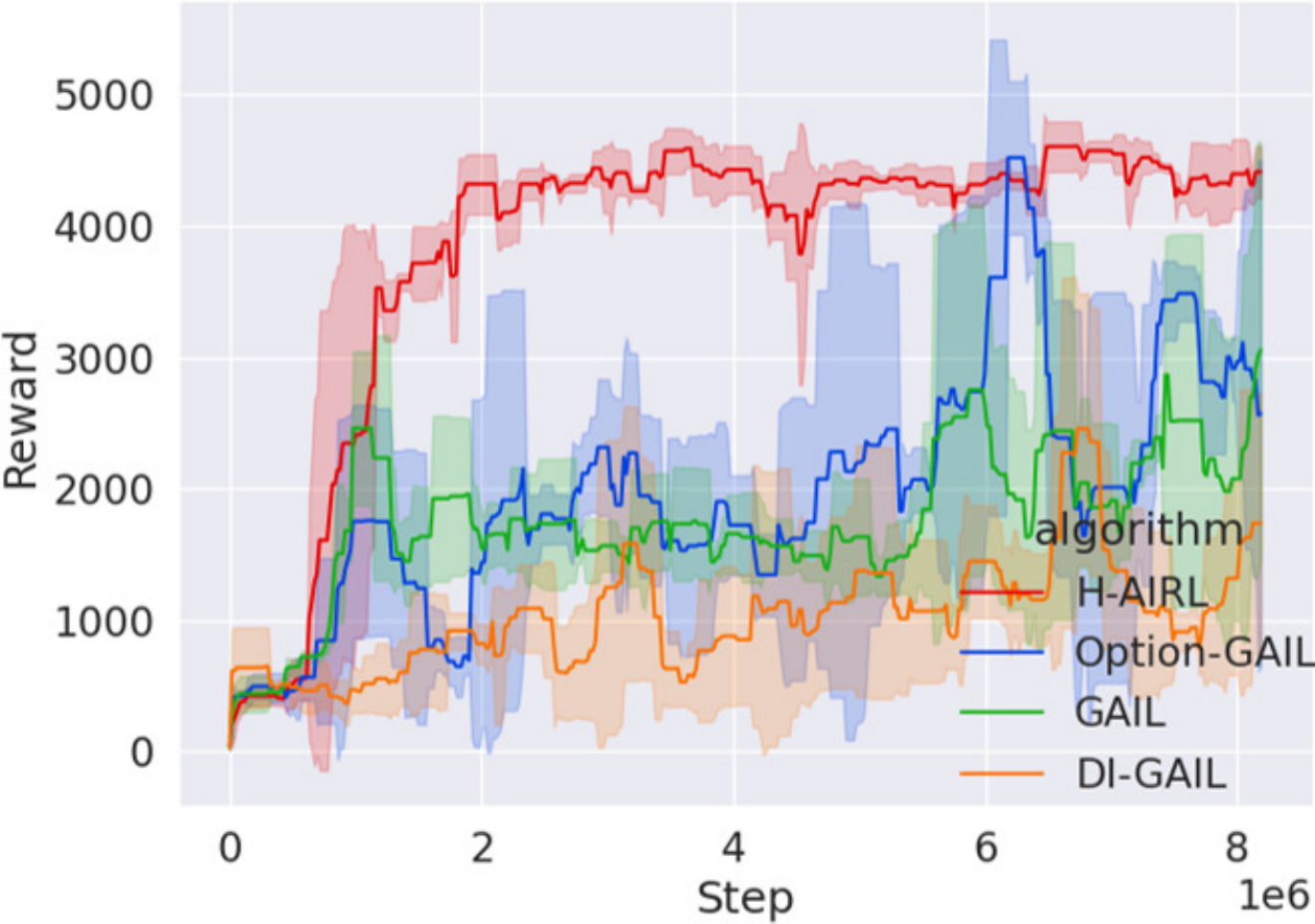}}
\subfigure[AntPusher]{
\label{fig:4(c)} 
\includegraphics[width=2.25in, height=1.3in]{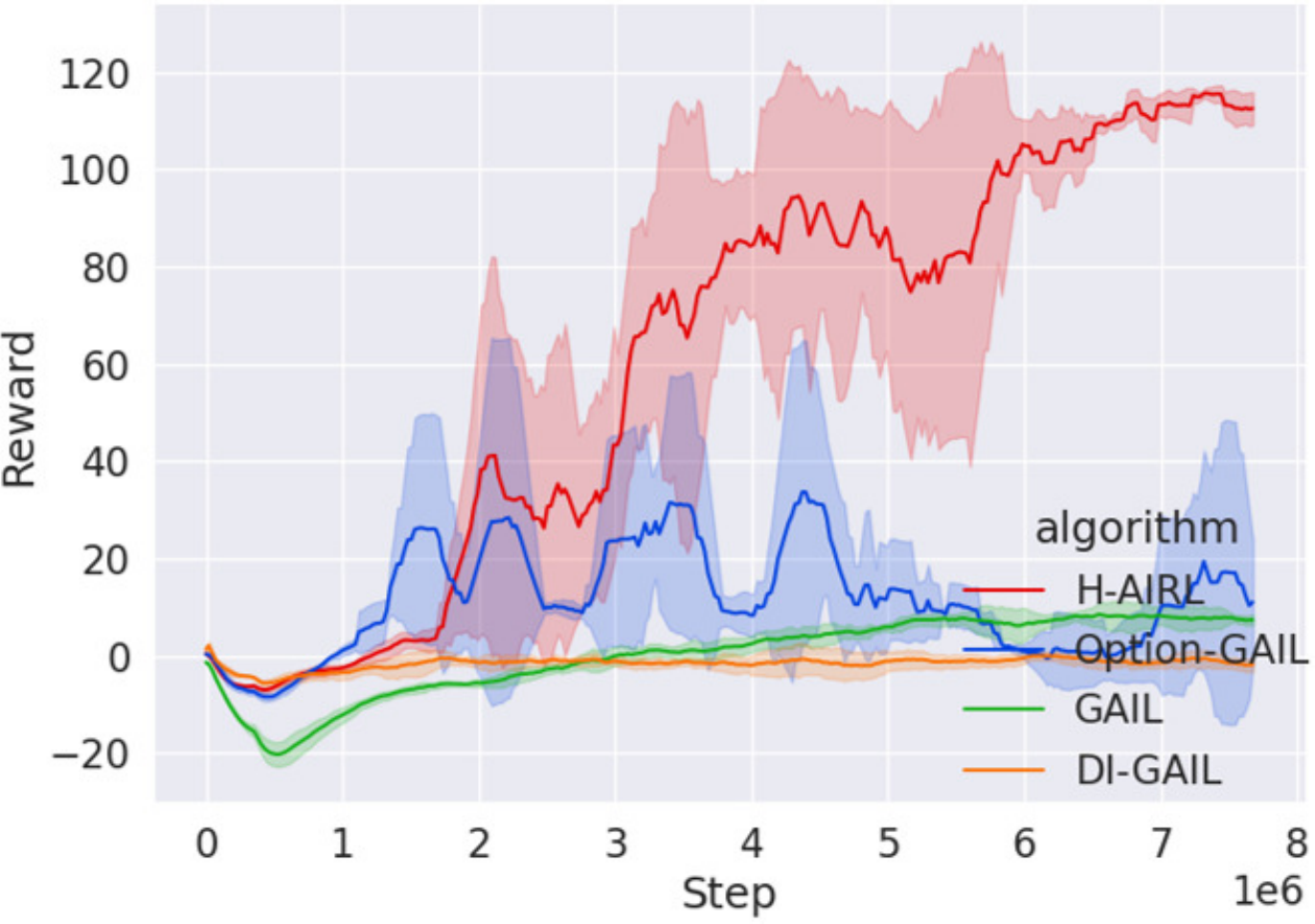}}
\caption{Comparisons with SOTA HIL and IL algorithms on Mujoco. Each training is repeated three times with different random seeds, and the mean and standard deviation are plotted as the solid lines and shaded areas, respectively. It can be observed that our algorithm, i.e., H-AIRL, performs the best, the GAIL-based algorithms suffer from unstableness, and the two-stage learning of the hierarchical policy in Directed-Info GAIL leads to poor performance.}
\label{fig:4} 
\end{figure*}

\subsection{Hierarchical Adversarial Inverse Reinforcement Learning} \label{h-airl}

Imitation Learning algorithms have been adopted to learn a policy from expert demonstrations rather than based on the reward signals which are usually difficult to acquire in real-life tasks. However, state-of-the-art IL algorithms like AIRL \cite{DBLP:journals/corr/abs-1710-11248} can not be directly adopted to recover a hierarchical policy since they don't take the local latent codes $Z_{0:T}$ into consideration.  Thus, we propose a novel hierarchical extension of AIRL, denoted as H-AIRL, as a solution, which is also part of our contribution. Further, we note that it is usually difficult to annotate the local latent codes $Z_{0:T}$ for an expert trajectory $X_{0:T}$, so we propose an Expectation-Maximization (EM) adaption of H-AIRL as well to learn the hierarchical policy based on only the unsegmented expert trajectories, i.e., $\{X_{0:T}\}$.

First, we give out the definition of the hierarchical policy. When observing a state $S_t$ at timestep $t \in \{0,\cdots, T-1\}$, the agent needs first to decide on its option choice based on $S_t$ and its previous option choice $Z_t$ using the high-level policy $\pi_{\theta}(Z_{t+1}|S_t,Z_t)$, and then decide on the action with the corresponding low-level policy $\pi_{\phi}(A_t|S_t,Z_{t+1})$. Thus, the hierarchical policy can be acquired with the chain rule as:
\begin{equation} \label{equ:7}
\begin{aligned}
    \pi_{\theta}(Z_{t+1}|S_t,Z_t)\cdot\pi_{\phi}(A_t|S_t,Z_{t+1})&=\pi_{\theta, \phi}(Z_{t+1},A_{t}|S_{t}, Z_{t}) \\
    &=\pi_{\theta, \phi}(\widetilde{A}_t|\widetilde{S}_t)    
\end{aligned}
\end{equation}
where the first equality holds because of the one-step Markov assumption, $\widetilde{S}_t \triangleq (S_t,Z_{t})$ and $\widetilde{A}_t \triangleq (Z_{t+1},A_{t})$ denote the extended state and action space respectively.

Next, by substituting $(S_t,A_t)$ with $(\widetilde{S}_t, \widetilde{A}_t)$ and $\tau_E$ with the hierarchical trajectory $(X_{0:T}, Z_{0:T})$ in Equation \eqref{equ:1}, we can get an MLE problem shown as Equation \eqref{equ:8}, from which we can recover the hierarchical reward function and policy. The derivation of \eqref{equ:8} is available in Appendix \ref{mle-obj}.
\begin{equation} \label{equ:8}
\begin{aligned}
    & \qquad\ \ \mathop{max}_{\vartheta}\mathbb{E}_{(X_{0:T}, Z_{0:T}) \sim \pi_{E}(\cdot)}\left[\log P_{\vartheta}(X_{0:T}, Z_{0:T})\right] \\
    &P_{\vartheta}(X_{0:T}, Z_{0:T})\propto \widehat{P}_{\vartheta}(X_{0:T}, Z_{0:T})\\ &=\mu(\widetilde{S}_0)\mathop{\prod}_{t=0}^{T-1}\mathcal{P}(\widetilde{S}_{t+1}|\widetilde{S}_{t}, \widetilde{A}_{t})\exp(\mathcal{R}_{\vartheta}(\widetilde{S}_t, \widetilde{A}_t)) \\
    &=\mu(S_0)\mathop{\prod}_{t=0}^{T-1}\mathcal{P}(S_{t+1}|S_t, A_{t})\exp(\mathcal{R}_{\vartheta}(S_t, Z_{t}, Z_{t+1}, A_{t}))\\
\end{aligned}
\end{equation}

As mentioned in Section \ref{airl}, the MLE problem can be efficiently solved with an adversarial learning framework, which is summarized as Equation \eqref{equ:10}. At optimality, we can recover the hierarchical reward function (i.e., $f_{\vartheta}$) and policy (i.e., $\pi_{\theta, \phi}$) of the expert with these objectives, of which the justification is provided in Appendix \ref{just}.
\begin{equation} \label{equ:10}
\begin{aligned}
        &\mathop{min}_{\vartheta}-\mathbb{E}_{(X_{0:T}, Z_{0:T}) \sim \pi_{E}(\cdot)}\left[\mathop{\sum}_{t=0}^{T-1}\log D_{\vartheta}(S_t,Z_t,Z_{t+1},A_t)\right] \\
        &- \mathbb{E}_{(X_{0:T}, Z_{0:T}) \sim \pi_{\theta, \phi}(\cdot)}\left[\mathop{\sum}_{t=0}^{T-1}\log(1-D_{\vartheta}(S_t,Z_t,Z_{t+1},A_t))\right] \\
        &\qquad\qquad\mathop{max}_{\theta, \phi}L^{IL}=\mathbb{E}_{(X_{0:T}, Z_{0:T}) \sim \pi_{\theta, \phi}(\cdot)}\mathop{\sum}_{t=0}^{T-1}R_{IL}^{t}
\end{aligned}
\end{equation}
In the above equation, the reward is defined as $R_{IL}^{t}=\log D_{\vartheta}^t - \log(1-D_{\vartheta}^t)$ and $D_{\vartheta}^t = D_{\vartheta}(S_t,Z_t,Z_{t+1},A_t)=\frac{\exp(f_{\vartheta}(S_t,Z_t,Z_{t+1},A_t))}{\exp(f_{\vartheta}(S_t,Z_t,Z_{t+1},A_t))+\pi_{\theta, \phi}(Z_{t+1},A_t|S_t,Z_t)}$.

Usually, we can only acquire the trajectories of state-action pairs, i.e., $X_{0:T}$, from the expert. In this case, we view the latent contexts $Z_{0:T}$ as hidden variables and adopt an EM-style adaption of our algorithm. In the expectation (E) step, we sample possible local latent codes with $Z_{0:T} \sim P_{\overline{\omega}}(\cdot|X_{0:T})$. $P_{\overline{\omega}}$ represents the trained posterior distribution network for $Z_{0:T}$, with the parameter $\overline{\omega}$, i.e., the old parameters before being updated in the M step. Then, in the maximization (M) step, we optimize the objectives shown in Equation \eqref{equ:10} for iterations, by replacing the samples in the first term of Equation \eqref{equ:10} with $(X_{0:T}, Z_{0:T})$ collected in the E step. The theoretical justification of the effectiveness of this EM-like algorithm is provided in Appendix \ref{em-airl}.

To sum up, there are in total four networks to learn in our system: the high-level policy $\pi_{\theta}$, low-level policy $\pi_{\phi}$, discriminator $D_{\vartheta}$, and variational posterior $P_{\omega}$. $D_{\vartheta}$ can be trained by minimizing the cross entropy loss shown in Equation \eqref{equ:10}. While, the update of the other three networks should follow:
\begin{equation} \label{equ:19}  
\begin{aligned}
   \mathop{max}_{\theta, \phi, \omega} L=\alpha_1 L^{DI}(\theta, \phi, \omega) + \alpha_2 L^{IL}(\theta, \phi)
\end{aligned}
\end{equation}
where $\alpha_{1:2}>0$ are the weights for each objective term and fine-tuned as hyperparameters, $L^{DI}$ and $L^{IL}$ are defined in Equation \eqref{equ:15} and \eqref{equ:10}, respectively. The pseudo code of our algorithm is available at Appendix \ref{pcode}.

%% file: evaluation.tex
\begin{table*}[t]
\centering
\caption{Numeric results of the ablation study}
\resizebox{\textwidth}{12mm}{
 \setlength{\tabcolsep}{7mm}{
\begin{tabular}{c || c c c c}
\hline
{ } & {Expert} & {H-AIRL (Ours)} & {Option-AIRL} & {H-GAIL} \\
\hline
\hline
{Hopper} & {$3139.85 \pm 712.02$} & {\bm{$3501.81 \pm 110.79$}} & {$1841.19 \pm 401.22$} & {$2574.06 \pm 920.34$}\\
\hline
{Walker} & {$5317.56 \pm 99.90$} & {\bm{$4354.14 \pm 193.28$}} & {$3951.41 \pm 631.48$} & {$3812.99 \pm 712.83$}\\
\hline
\makecell[c]{AntPusher}
 & {$116.00 \pm 3.99$} & {\bm{$113.94 \pm 2.54$}} & {$81.58 \pm 39.75$} & {$80.23 \pm 30.29$} \\  
\hline
\end{tabular}}}
\label{table:1}
\end{table*}

\section{EVALUATION AND MAIN RESULTS}

In this section, we compare H-AIRL with SOTA HIL algorithms: Option-GAIL \cite{DBLP:conf/icml/JingH0MKGL21} and Directed-Info GAIL \cite{DBLP:conf/iclr/SharmaSRK19}, to justify the superiority of our algorithm, and we provide comparisons with SOTA IL algorithms: GAIL \cite{ho2016generative} to show the importance of hierarchical policy learning for challenging long-horizon tasks. To keep it fair, we use the original implementations of these baseline algorithms provided by the authors. Moreover, we provide ablation study of our algorithm to evaluate the key components of our algorithm design. Specifically, we compare with (i) Option-AIRL: a version of our algorithm by only keeping the AIRL-related term in the objective to update the hierarchical policy, i.e., $L^{IL}$ in Equation \eqref{equ:19}; (ii) H-GAIL: a variant by replacing the AIRL component of our algorithm with GAIL, of which the details are in Appendix \ref{h-gail}.

As shown in Figure \ref{fig:3}, these algorithms are evaluated on three challenging continuous control robotic tasks built with Mujoco \cite{todorov2012mujoco}, i.e., Hopper, Walker and AntPusher. Hopper and Walker are locomotion tasks where the robot agents are required to move toward a certain direction by learning to coordinate their legs. Both of them have continuous state and action spaces. Specifically, Hopper has a 11-dim state space and 3-dim action space, and Walker has a 17-dim state space and 6-dim action space. While, the AntPusher needs not only to learn locomotion skills for the Ant agent (Figure \ref{fig:3(c)}) but also to learn to navigate into a room that is blocked by a movable box (i.e., the red one in Figure \ref{fig:3(d)}). In particular, the Ant agent needs to first navigate to the left side of the box and push it away, and then enter the blocked room to complete the task, which is much more challenging with a 107-dim continuous state space and 8-dim continuous action space. Note that all of the three tasks are long-horizon episodic tasks with a horizon of 1000 time steps. The expert demonstrations for imitation are generated from well-trained policies by PPO \cite{DBLP:journals/corr/SchulmanWDRK17} with $2\times10^7$ exploration steps.

\subsection{Main Results}

First, we compare our algorithm, i.e., H-AIRL, with the SOTA HIL and IL baselines mentioned above on the three Mujoco tasks. As shown in Figure \ref{fig:4}, we plot the change of the episodic rewards (i.e., the sum of the rewards at each time step within an episode) in the training process. Note that these episodic rewards are specifically designed by OpenAI Gym \cite{DBLP:journals/corr/BrockmanCPSSTZ16} to encourage the Mujoco agents to complete the corresponding tasks as fast as possible at the least control cost, which can be used as evaluation metrics for the agents' learning performance. We repeat each training for three times with different random seeds, plot the average value as the solid lines and the standard deviation as the shaded areas. It can be observed from Figure \ref{fig:4} that H-AIRL outperforms the baselines significantly in terms of both the convergence speed and final performance. Also, Option-GAIL has better performance than GAIL, which shows the advantages of hierarchical policy learning for challenging long-horizon tasks. While, the fluctuations during the training process of GAIL and Option-GAIL show the unstableness of the GAIL-based algorithms. Moreover, it can be observed that Directed-Info GAIL performs even worse than GAIL which does not take advantage of options in the learning process, showing that the separate learning of the high-level and low-level policy (i.e., learning the high-level policy first at the pretraining stage and then fixing it during the low-level policy training) will harm the agents' performance and lead to convergence to a poor local optimum.

Next, we provide comparisons of H-AIRL with Option-AIRL and H-GAIL as the ablation study. As shown in Equation \eqref{equ:19}, the objective function for updating the hierarchical policy includes two parts, i.e., the directed information term $L^{DI}$ and AIRL term $L^{IL}$. In order to evaluate the importance of $L^{DI}$, we implement the baseline Option-AIRL, for which we only keep $L^{IL}$, i.e., the AIRL objectives on the extended state and action spaces, for updating the hierarchical policy. On the other hand, we replace the AIRL objective with the one defined with GAIL (Equation \eqref{equ:a35}), denoted as H-GAIL, to show the necessity to adopt AIRL as our base imitation learning algorithm. In Table \ref{table:1}, we provide the numeric results of the performance of the expert demonstrations, our algorithm, and the ablation baselines on the three Mujoco benchmarks. To be specific, we repeat the training with each algorithm on each task for three times with different random seeds, and calculate the mean and standard deviation of the episodic rewards after they converge across different random seeds as the metric of their final performance. It can be observed that our algorithm outperforms the baselines in both the average performance and the stableness, showing the effectiveness of the key components of our algorithm design. On the other hand, even the ablation variants of our algorithm, i.e., Option-AIRL and H-GAIL, have better final performance than the baselines shown in Figure \ref{fig:4}, which further shows the superiority of our algorithm.

%% file: conclusion.tex
\section{CONCLUSION}

Hierarchical Imitation Learning has been proved to outperform canonical Imitation Learning when the demonstrated task is complex and has a subtask structure. In this paper, we propose a novel HIL algorithm by integrating the extended AIRL objectives with a directed information term, and provide a VAE-like framework for updating the target hierarchical policy in an end-to-end fashion. Further, we propose an EM adaption of our algorithm to fit it with unsegmented expert demonstrations to allow more practicability. We also have provided theoretical analysis and ablation study of each key component of our algorithm design for justification, and comparisons on Mujoco control tasks with SOTA HIL and IL baselines to show the superiority of our algorithm.

As for future works, applying H-AIRL to realistic robotic tasks based on human demonstrations can be a solid practical contribution. Integrating H-AIRL with Meta/Multi-task Learning techniques \cite{DBLP:conf/corl/FinnYZAL17, DBLP:conf/icra/DeisenrothEPF14, DBLP:conf/icra/SinghJIKDLKF20} for novel Multi-task HIL algorithms is also an interesting direction.

%% file: appendix.tex
\appendices
\onecolumn

\section{Implementation of the Hierarchical Policy in the One-step Option Model} \label{mha}

In this section, we give out the detailed structure design of the hierarchical policy introduced in Section \ref{option}, i.e., $\pi_{\theta}(Z|S, Z')$ and $\pi_{\phi}(A|S,Z)$, which is proposed in \cite{li2020skill}. This part is not our contribution, so we only provide the details for the purpose of implementation.

As mentioned in Section \ref{option}, the structure design is based on the  Multi-Head Attention (MHA) mechanism \cite{DBLP:conf/nips/VaswaniSPUJGKP17}. An attention function can be described as mapping a query, i.e., $q \in \mathbb{R}^{d_k}$, and a set of key-value pairs, i.e., $K=[k_1 \cdots k_n]^T \in \mathbb{R}^{n \times d_k}$ and $V=[v_1 \cdots v_n]^T \in \mathbb{R}^{n \times d_v}$, to an output. The output is computed as a weighted sum of the values, where the weight assigned to each value is computed by a compatibility function of the query with the corresponding key. To be specific:
\begin{equation} \label{equ:a8}  
\begin{aligned}
    Attention(q, K, V) = \mathop{\sum}_{i=1}^n\left[\frac{exp(q \cdot k_i)}{\mathop{\sum}_{j=1}^n exp(q \cdot k_j)} \times v_i \right]
\end{aligned}
\end{equation}
where $q,K,V$ are learnable parameters, $\frac{exp(q \cdot k_i)}{\mathop{\sum}_{j=1}^n exp(q \cdot k_j)}$ represents the attention weight that the model should pay to item $i$. In MHA, the query and key-value pairs are first linearly projected $h$ times to get $h$ different queries, keys and values. Then, an attention function is performed on each of these projected versions of queries, keys and values in parallel to get $h$ outputs which are then be concatenated and linearly projected to acquire the final output. The whole process can be represented as Equation \eqref{equ:a9}, where $W_i^q \in \mathbb{R}^{d_k \times d_k}, W_i^K \in \mathbb{R}^{d_k \times d_k}, W_i^V \in \mathbb{R}^{d_v \times d_v}, W^O \in \mathbb{R}^{nd_v \times d_v}$ are the learnable parameters. 
\begin{equation} \label{equ:a9}  
\begin{aligned}
    MHA(q,K,V)=Concat(head_1, \cdots, head_h)W^O,\  head_i=Attention(qW_i^q, KW_i^K, VW_i^V)
\end{aligned}
\end{equation}

In this work, the option is represented as an $N$-dimensional one-hot vector, where $N$ denotes the total number of options to learn. The high-level policy $\pi_{\theta}(Z|S, Z')$ has the structure shown as:
\begin{equation} \label{equ:a10}  
\begin{aligned}
    q=linear(Concat[S, W_C^TZ']),\ dense_Z = MHA(q, W_C, W_C),\ Z \sim Categorical(\cdot|dense_Z)
\end{aligned}
\end{equation}
$W_C \in \mathbb{R}^{N \times E}$ is the option context matrix of which the $i$-th row represents the context embedding of the option $i$. $W_C$ is also used as the key and value matrix for the MHA, so $d_k=d_v=E$ in this case. Note that $W_C$ is only updated in the MHA module. Intuitively, $\pi_{\theta}(Z|S, Z')$ attends to all the option context embeddings in $W_C$ according to $S$ and $Z'$. If $Z'$ still fits $S$, $\pi_{\theta}(Z|S, Z')$ will assign a larger attention weight to $Z'$ and thus has a tendency to continue with it; otherwise, a new skill with better compatibility will be sampled.

As for the low-level policy $\pi_{\phi}(A|S,Z)$, it has the following structure:
\begin{equation} \label{equ:a11}  
\begin{aligned}
    dense_A = MLP(S, W_C^TZ),\ A \sim Categorical/Gaussian(\cdot|dense_A)
\end{aligned}
\end{equation}
where $MLP$ represents a multilayer perceptron, $A$ follows a categorical distribution for the discrete case or a gaussian distribution for the continuous case. The context embedding corresponding to $Z$, i.e., $W_C^TZ$, instead of $Z$ only, is used as input of $\pi_{\phi}$ since it can encode multiple properties of the option $Z$ \cite{DBLP:conf/nips/KosiorekSTH19}. 

\section{A Lower Bound of the Directed Information Objective} \label{dilb}

In this section, we give out the derivation of a lower bound of the directed information from the trajectory sequence $X_{0:T}$ to the local latent context sequence $Z_{0:T}$, i.e., $I(X_{0:T} \rightarrow Z_{0:T})$ as follows:
\begin{equation} \label{equ:a1}  
\begin{aligned}
    &I(X_{0:T} \rightarrow Z_{0:T}) =  \mathop{\sum}_{t=1}^{T}\left[I(X_{0:t};Z_{t}|Z_{0:t-1})\right] \\
    &= \mathop{\sum}_{t=1}^{T}\left[H(Z_t|Z_{0:t-1})-H(Z_t|X_{0:t},Z_{0:t-1})\right] \\
    &\geq \mathop{\sum}_{t=1}^{T}\left[H(Z_t|X_{0:t-1}, Z_{0:t-1})-H(Z_t|X_{0:t},Z_{0:t-1})\right] \\
    &= \mathop{\sum}_{t=1}^{T} [H(Z_t|X_{0:t-1}, Z_{0:t-1}) +  \mathop{\sum}_{\substack{X_{0:t},\\ Z_{0:t-1}}}P(X_{0:t}, Z_{0:t-1})\mathop{\sum}_{Z_t}P(Z_t|X_{0:t}, Z_{0:t-1})logP(Z_t|X_{0:t}, Z_{0:t-1})]
\end{aligned}
\end{equation}
In Equation \eqref{equ:a1}, $I(Var_1;Var_2|Var_3)$ denotes the conditional mutual information, $H(Var_1|Var_2)$ denotes the conditional entropy, and the inequality holds because of the basic property related to conditional entropy: increasing conditioning cannot increase entropy \cite{galvin2014three}. $H(Z_t|X_{0:t-1}, Z_{0:t-1})$ is the entropy of the high-level policy (introduced later), which can serve as a regulator and more convenient to obtain. Further, the second term in the last step can be processed as follows:
\begin{equation} \label{equ:a2}  
\begin{aligned}
    &\mathop{\sum}_{Z_t}P(Z_t|X_{0:t}, Z_{0:t-1})logP(Z_t|X_{0:t}, Z_{0:t-1})\\
    &= \mathop{\sum}_{Z_t}P(Z_t|X_{0:t}, Z_{0:t-1})\left[log\frac{P(Z_t|X_{0:t}, Z_{0:t-1})}{P_{\omega}(Z_t|X_{0:t}, Z_{0:t-1})} + logP_{\omega}(Z_t|X_{0:t}, Z_{0:t-1})\right]\\
    &= D_{KL}(P(\cdot|X_{0:t}, Z_{0:t-1})||P_{\omega}(\cdot|X_{0:t}, Z_{0:t-1})) + \mathop{\sum}_{Z_t}P(Z_t|X_{0:t}, Z_{0:t-1})logP_{\omega}(Z_t|X_{0:t}, Z_{0:t-1})\\
    &\geq \mathop{\sum}_{Z_t}P(Z_t|X_{0:t}, Z_{0:t-1})logP_{\omega}(Z_t|X_{0:t}, Z_{0:t-1})
\end{aligned}
\end{equation}
where $D_{KL}(\cdot)$ denotes the Kullback-Leibler (KL) Divergence which is non-negative \cite{cover1999elements}, $P_{\omega}(Z_t|X_{0:t}, Z_{0:t-1})$ is a variational estimation of the posterior distribution of $Z_t$ given $X_{0:t}$ and $Z_{0:t-1}$, i.e., $P(Z_t|X_{0:t}, Z_{0:t-1})$, which is modeled as a recurrent neural network with the parameter set $\omega$ in our work. Based on Equation \eqref{equ:a1} and \eqref{equ:a2}, we can obtain a lower bound of $I(X_{0:T} \rightarrow Z_{0:T})$ denoted as $L^{DI}$:
\begin{equation} \label{equ:a3}  
\begin{aligned}
    L^{DI}=\mathop{\sum}_{t=1}^{T}[\mathop{\sum}_{\substack{X_{0:t}, Z_{0:t}}}P(X_{0:t}, Z_{0:t})logP_{\omega}(Z_t|X_{0:t}, Z_{0:t-1}) + H(Z_t|X_{0:t-1}, Z_{0:t-1})]
\end{aligned}
\end{equation}
Note that the joint distribution $P(X_{0:t}, Z_{0:t})$ has a recursive definition as follows:
\begin{equation} \label{equ:a4}  
\begin{aligned}
    &P(X_{0:t}, Z_{0:t}) = P(X_{t}|X_{0:t-1}, Z_{0:t})P(Z_{t}|X_{0:t-1}, Z_{0:t-1})P(X_{0:t-1}, Z_{0:t-1}) \\
    & \qquad \qquad \qquad P(X_{0}, Z_{0}) = P((S_{0},A_{-1}), Z_{0}) = \mu(S_{0})
\end{aligned}
\end{equation}
where $\mu(S_{0})$ denotes the distribution of the initial states. The second line in Equation \eqref{equ:a4} holds because $A_{-1}$ and $Z_{0}$ are dummy variables which are only for keeping the notations consistent and never executed and set to be constant. Based on Equation \eqref{equ:a4}, we can get:
\begin{equation} \label{equ:a6}  
\begin{aligned}
    &P(X_{0:t}, Z_{0:t}) = \mu(S_{0})\mathop{\prod}_{i=1}^{t}P(Z_{i}|X_{0:i-1}, Z_{0:i-1})P(X_{i}|X_{0:i-1}, Z_{0:i})\\
    &=\mu(S_{0})\mathop{\prod}_{i=1}^{t}P(Z_{i}|X_{0:i-1}, Z_{0:i-1})P((S_i, A_{i-1})|X_{0:i-1}, Z_{0:i})\\
    &=\mu(S_{0})\mathop{\prod}_{i=1}^{t}P(Z_{i}|X_{0:i-1}, Z_{0:i-1})P( A_{i-1}|X_{0:i-1}, Z_{0:i})\mathcal{P}(S_i|S_{i-1},A_{i-1})
\end{aligned}
\end{equation}
In Equation \eqref{equ:a6}, $\mathcal{P}(S_i|S_{i-1},A_{i-1})$ is the transition dynamic, $P(Z_{i}|X_{0:i-1}, Z_{0:i-1})$ is output of the high-level policy which decides on the current option $Z_{i}$ based on the history trajectory $X_{0:i-1}$ and option choices $Z_{0:i-1}$, $P( A_{i-1}|X_{0:i-1}, Z_{0:i})$ is output of the low-level policy which outputs the action choice based on the history information and current option choice.

To sum up, we can adopt the high-level policy, low-level policy and variational posterior to get an estimation of the lower bound of the directed information objective through Monte Carlo sampling \cite{sutton2018reinforcement} according to Equation \eqref{equ:a3} and \eqref{equ:a6}.

\section{Implementation Details of the Variational Posteriors} \label{rnn-imp}

The variational posterior for the local latent code, i.e., $P_{\omega}(Z_t|X_{0:t}, Z_{0:t-1})$, is modeled as $P_{\omega}(Z_t|X_t, Z_{t-1}, h_{t-1})$, where $h_{t-1}$ is the internal hidden state of an RNN. $h_{t-1}$ is recursively maintained with the time series using the GRU rule, i.e., $h_{t-1}=GRU(X_{t-1},Z_{t-2},h_{t-2})$, to embed the history information in the trajectory, i.e., $X_{0:t-1}$ and $Z_{0:t-2}$. Note that the RNN-based posterior has been used and justified in the process for sequential data \cite{DBLP:conf/nips/ChungKDGCB15}.

\section{Derivation of Equation (\ref{equ:8}) in the MLE Objective} \label{mle-obj}
In Equation \eqref{equ:a31}, $Z_0$ is a dummy variable which is assigned before the episode begins and never executed. It's implemented as a constant across different episodes, so we have $P(S_0,Z_0)=P(S_0)=\mu(S_0)$, where $\mu(\cdot)$ denotes the initial state distribution. Also, we have $P(S_{t+1},Z_{t+1}|S_t, Z_t, Z_{t+1}, A_{t})=P(Z_{t+1}|S_t, Z_t, Z_{t+1}, A_{t})P(S_{t+1}|S_t, Z_t, Z_{t+1}, A_{t})=\mathcal{P}(S_{t+1}|S_t, A_{t})$, since the transition dynamic $\mathcal{P}$ is irrelevant to the local latent codes $Z$ based on the definition of MDP.
\begin{equation} \label{equ:a31}
\begin{aligned}
&P_{\vartheta}(X_{0:T}, Z_{0:T})\propto \mu(\widetilde{S}_0)\mathop{\prod}_{t=0}^{T-1}\mathcal{P}(\widetilde{S}_{t+1}|\widetilde{S}_{t}, \widetilde{A}_{t})exp(\mathcal{R}_{\vartheta}(\widetilde{S}_t, \widetilde{A}_t)) \\
&=P(S_0,Z_0)\mathop{\prod}_{t=0}^{T-1}P(S_{t+1},Z_{t+1}|S_t, Z_t, Z_{t+1}, A_{t})exp(\mathcal{R}_{\vartheta}(S_t, Z_{t}, Z_{t+1}, A_{t})\\
&=\mu(S_0)\mathop{\prod}_{t=0}^{T-1}\mathcal{P}(S_{t+1}|S_t, A_{t})exp(\mathcal{R}_{\vartheta}(S_t, Z_{t}, Z_{t+1}, A_{t}))\\
\end{aligned}
\end{equation}

\section{Justification of the Objective Function Design of H-AIRL in Equation (\ref{equ:10})} \label{just}

In this section, we prove that by optimizing the objective functions shown as Equation \eqref{equ:10}, we can get the solution of the MLE problem shown as Equation \eqref{equ:8}, i.e., the hierarchical reward function and policy of the expert.

In Appendix A of \cite{DBLP:journals/corr/abs-1710-11248}, they show that the discriminator objective is equivalent to the MLE objective where $f_{\vartheta}$ serves as $R_{\vartheta}$, when $D_{KL}(\pi(\tau)||\pi_E(\tau))$ is minimized. The same conclusion can be acquired by simply replacing $\{S_t,A_t,\tau\}$ with $\{(S_t, Z_t),(Z_{t+1}, A_t),(X_{0:T}, Z_{0:T})\}$, i.e., the extended definition of the state, action and trajectory, in the original proof, which we don't repeat here. Then, we only need to prove that $D_{KL}(\pi_{\theta, \phi}(X_{0:T}, Z_{0:T})||\pi_E(X_{0:T}, Z_{0:T}$)) can be minimized through Equation \eqref{equ:10}:
\begin{equation} \label{equ:a32}
\begin{aligned}
    &\mathop{max}_{\theta, \phi}\mathbb{E}_{(X_{0:T}, Z_{0:T}) \sim \pi_{\theta, \phi}(\cdot)}\mathop{\sum}_{t=0}^{T-1}R_{IL}^{t}\\
    &=\mathop{\mathbb{E}}_{X_{0:T}, Z_{0:T}}\left[\mathop{\sum}_{t=0}^{T-1}logD_{\vartheta}(S_t,Z_t,Z_{t+1},A_t) - log(1-D_{\vartheta}(S_t,Z_t,Z_{t+1},A_t))\right]\\
    &=\mathop{\mathbb{E}}_{X_{0:T}, Z_{0:T}}\left[\mathop{\sum}_{t=0}^{T-1}f_{\vartheta}(S_t,Z_t,Z_{t+1},A_t) - log\pi_{\theta, \phi}(Z_{t+1},A_t|S_t,Z_t)\right]\\
    &=\mathop{\mathbb{E}}_{X_{0:T}, Z_{0:T}}\left[\mathop{\sum}_{t=0}^{T-1}f_{\vartheta}(S_t,Z_t,Z_{t+1},A_t) - log(\pi_{\theta}(Z_{t+1}|S_t,Z_t)\pi_{\phi}(A_t|S_t,Z_{t+1}))\right]\\
    &=\mathop{\mathbb{E}}_{X_{0:T}, Z_{0:T}}\left[log\frac{\prod_{t=0}^{T-1}exp(f_{\vartheta}(S_t,Z_t,Z_{t+1},A_t))}{\prod_{t=0}^{T-1}\pi_{\theta}(Z_{t+1}|S_t,Z_t)\pi_{\phi}(A_t|S_t,Z_{t+1})}\right]\\
    &\iff\mathop{max}_{\theta, \phi}\mathop{\mathbb{E}}_{X_{0:T}, Z_{0:T}}\left[log\frac{\prod_{t=0}^{T-1}exp(f_{\vartheta}(S_t,Z_t,Z_{t+1},A_t))/Z_{\vartheta}}{\prod_{t=0}^{T-1}\pi_{\theta}(Z_{t+1}|S_t,Z_t)\pi_{\phi}(A_t|S_t,Z_{t+1})}\right]\\
\end{aligned}
\end{equation}
Note that $Z_{\vartheta}=\sum_{X_{0:T}, Z_{0:T}}\widehat{P}_{\vartheta}(X_{0:T}, Z_{0:T})$ (defined in Equation \eqref{equ:8}) is the normalized function parameterized with $\vartheta$, so the introduction of $Z_{\vartheta}$ will not influence the optimization with respect to $\theta$ and $\phi$ and the equivalence at the last step holds. Also, the second equality shows that the hierarchical policy is recovered by optimizing an entropy-regularized policy objective where $f_{\vartheta}$ serves as $R_{\vartheta}$. Further, we have:
\begin{equation} \label{equ:a33}
\begin{aligned}
    &\mathop{max}_{\theta, \phi}\mathop{\mathbb{E}}_{X_{0:T}, Z_{0:T}}\left[log\frac{\prod_{t=0}^{T-1}exp(f_{\vartheta}(S_t,Z_t,Z_{t+1},A_t))/Z_{\vartheta}}{\prod_{t=0}^{T-1}\pi_{\theta}(Z_{t+1}|S_t,Z_t)\pi_{\phi}(A_t|S_t,Z_{t+1})}\right]\\
    &=\mathop{\mathbb{E}}_{X_{0:T}, Z_{0:T}}\left[log\frac{\mu(S_0)\prod_{t=0}^{T-1}\mathcal{P}(S_{t+1}|S_t, A_{t})\prod_{t=0}^{T-1}exp(f_{\vartheta}(S_t,Z_t,Z_{t+1},A_t))/Z_{\vartheta}}{\mu(S_0)\prod_{t=0}^{T-1}\mathcal{P}(S_{t+1}|S_t, A_{t})\prod_{t=0}^{T-1}\pi_{\theta}(Z_{t+1}|S_t,Z_t)\pi_{\phi}(A_t|S_t,Z_{t+1})}\right]\\
    &=\mathbb{E}_{(X_{0:T}, Z_{0:T}) \sim \pi_{\theta, \phi}(\cdot)}\left[log\frac{\pi_{E}(X_{0:T},Z_{0:T})}{\pi_{\theta, \phi}(X_{0:T},Z_{0:T})}\right]\\
    &=-D_{KL}(\pi_{\theta, \phi}(X_{0:T}, Z_{0:T})||\pi_E(X_{0:T}, Z_{0:T}))\\
    & \iff\mathop{min}_{\theta, \phi}D_{KL}(\pi_{\theta, \phi}(X_{0:T}, Z_{0:T})||\pi_E(X_{0:T}, Z_{0:T}))
\end{aligned}
\end{equation}
where the second equality holds because of the definition of $\pi_{E}$ (Equation \eqref{equ:8} with $f_{\vartheta}$ serving as $R_{\vartheta}$) and $\pi_{\theta, \phi}$ (Equation \eqref{equ:7}).

\section{Justification of the EM-style Adaption} \label{em-airl}

Given only a dataset of expert trajectories, i.e., $D_{E} \triangleq \{X_{0:T}\}$, we can still maximize the likelihood estimation $\mathbb{E}_{X_{0:T} \sim D_{E}}\left[logP_{\vartheta}(X_{0:T})\right]$ through an EM-style adaption:
\begin{equation} \label{equ:a34}
\begin{aligned}
    &\mathbb{E}_{X_{0:T} \sim D_{E}}\left[logP_{\vartheta}(X_{0:T})\right]=\mathbb{E}_{X_{0:T} \sim D_{E}}\left[log\left[\mathop{\sum}_{Z_{0:T}}P_{\vartheta}(X_{0:T},Z_{0:T})\right]\right]\\
    &=\mathbb{E}_{X_{0:T} \sim D_{E}}\left[log\left[\mathop{\sum}_{ Z_{0:T}}\frac{P_{\vartheta}(X_{0:T},Z_{0:T})}{P_{\overline{\vartheta}}(Z_{0:T}|X_{0:T})}P_{\overline{\vartheta}}(Z_{0:T}|X_{0:T})\right]\right]\\
    &=\mathbb{E}_{X_{0:T} \sim D_{E}}\left[log\left[\mathbb{E}_{Z_{0:T}\sim P_{\overline{\vartheta}}(\cdot|X_{0:T})}\frac{P_{\vartheta}(X_{0:T},Z_{0:T})}{P_{\overline{\vartheta}}(Z_{0:T}|X_{0:T})}\right]\right]\\
    &\geq \mathbb{E}_{X_{0:T} \sim D_{E}}\left[\mathbb{E}_{Z_{0:T}\sim P_{\overline{\vartheta}}(\cdot|X_{0:T})}log\frac{P_{\vartheta}(X_{0:T},Z_{0:T})}{P_{\overline{\vartheta}}(Z_{0:T}|X_{0:T})}\right]\\
    &= \mathbb{E}_{X_{0:T} \sim D_{E}, Z_{0:T} \sim P_{\overline{\omega}}(\cdot|X_{0:T})}\left[log\frac{P_{\vartheta}(X_{0:T},Z_{0:T})}{P_{\overline{\vartheta}}(Z_{0:T}|X_{0:T})}\right]\\
    &= \mathbb{E}_{X_{0:T}, Z_{0:T}}\left[logP_{\vartheta}(X_{0:T},Z_{0:T})\right] - \mathbb{E}_{X_{0:T}, Z_{0:T}}\left[logP_{\overline{\vartheta}}(Z_{0:T}|X_{0:T})\right]\\
\end{aligned}
\end{equation}
where we adopt the Jensen's inequality \cite{jensen1906fonctions} in the 4-th step. Also, we note that $P_{\overline{\omega}}(Z_{0:T}|X_{0_T})$ provides a posterior distribution of $Z_{0:T}$, which corresponds to the generating process led by the hierarchical policy. As justified in \ref{just}, the hierarchical policy is trained with the reward function parameterized with $\overline{\vartheta}$. Thus, the hierarchical policy is a function of $\overline{\vartheta}$, and the network $P_{\overline{\omega}}$ corresponding to the hierarchical policy provides a posterior distribution related to the parameter set $\overline{\vartheta}$, i.e., $Z_{0:T}\sim P_{\overline{\vartheta}}(\cdot|X_{0:T}) \iff  Z_{0:T} \sim P_{\overline{\omega}}(\cdot|X_{0:T})$, due to which the 5-th step holds. Note that $\overline{\vartheta}, \overline{\omega}$ denote the parameters $\vartheta, \omega$ before being updated in the M step.

In the second equality of Equation \eqref{equ:a34}, we introduce the sampled local latent codes in the E step as discussed in Section \ref{h-airl}. Then, in the M step, we optimize the objectives shown in Equation \eqref{equ:10} for iterations, by replacing the samples in the first term of Equation \eqref{equ:10} with $(X_{0:T}, Z_{0:T})$ collected in the E step.  This is equivalent to solve the MLE problem: $\mathop{max}_{\vartheta}\mathbb{E}_{X_{0:T} \sim D_{E}, Z_{0:T}\sim P_{\overline{\omega}}(\cdot|X_{0:T})}\left[logP_{\vartheta}(X_{0:T},Z_{0:T})\right]$, through which we can maximize a lower bound of the original objective, i.e., $\mathop{max}_{\vartheta}\mathbb{E}_{X_{0:T} \sim D_{E}}\left[logP_{\vartheta}(X_{0:T})\right]$, as shown in the last step of Equation \eqref{equ:a34}. Thus, the original objective can be optimized through this EM procedure. Note that the second term in the last step is a function of the old parameter $\overline{\vartheta}$ so that it can be overlooked when optimizing with respect to $\vartheta$.

\section{Pseudo Code of the Overall Algorithm} \label{pcode}

 \begin{algorithm}[t]
	\caption{Hierarchical Adversarial Inverse Reinforcement Learning (H-AIRL)}\label{alg:1}
	\begin{algorithmic}[1]
	    \State \textbf{Input:} Expert demonstrations $\{X_{0:T}^E\}$ (If the option annotations, i.e., $\{Z_{0:T}^E\}$, are provided, step 5 is not required.)
	    \State Initialize the posterior network $P_{\omega}$, high-level policy $\pi_{\theta}$, low-level policy $\pi_{\phi}$, and discriminator $D_{\vartheta}$
		\For {$each\ training\ episode$}
		    \State Generate $M$ trajectories $\{(X_{0:T}, Z_{0:T})\}$ with $\pi_{\theta}$ and $\pi_{\phi}$ by interacting with the simulator
		    \State Sample local latent codes corresponding to the expert trajectories using the posterior, i.e., $Z_{0:T}^E \sim P_{\omega}(\cdot|X_{0:T}^E)$
		    \State Update $P_{\omega}$ by minimizing $L^{DI}$ (Equation \eqref{equ:15}) using Stochastic Gradient Descent \cite{bottou2010large} with $\{(X_{0:T}, Z_{0:T})\}$
			\State Update $D_{\vartheta}$ by minimizing the cross entropy loss in Equation \eqref{equ:10} based on $\{(X_{0:T}, Z_{0:T})\}$ and $\{(X_{0:T}^E, Z_{0:T}^E)\}$
			\State Train $\pi_{\theta}$ and $\pi_{\phi}$ by maximizing the return function $L$ defined in Equation \eqref{equ:19} using the HRL algorithm SA \cite{li2020skill}
		\EndFor
	\end{algorithmic} 
\end{algorithm}

The pseudo code of H-AIRL is provided as Algorithm \ref{alg:1}.

\section{Implementation Details of H-GAIL} \label{h-gail}

H-GAIL is a variant of our algorithm by replacing the AIRL component with GAIL. Similar with Section \ref{h-airl}, we need to provide an extension of GAIL with the one-step option model, in order to learn a hierarchical policy. The extension method follows Option-GAIL \cite{DBLP:conf/icml/JingH0MKGL21} which is one of our baselines. H-GAIL also uses an adversarial learning framework that contains a discriminator $D_{\psi}$ and a hierarchical policy $\pi_{\theta, \phi}$, for which the objective are as follows:
\begin{equation} \label{equ:a35}
\begin{aligned}
        &\mathop{max}_{\psi}\mathbb{E}_{(S, A, Z, Z') \sim \pi_{E}(\cdot)}\left[log(1-D_{\psi}(S, A, Z, Z'))\right] + \mathbb{E}_{(S, A, Z, Z') \sim \pi_{\theta, \phi}(\cdot)}\left[logD_{\psi}(S, A, Z, Z')\right] \\
        &\qquad\qquad\mathop{max}_{\theta, \phi}L^{IL}=\mathbb{E}_{(X_{0:T}, Z_{0:T}) \sim \pi_{\theta, \phi}(\cdot)}\mathop{\sum}_{t=0}^{T-1}\left[-logD_{\psi}(S_t, A_t, Z_{t+1}, Z_t)\right]
\end{aligned}
\end{equation}
where $(S, A, Z, Z')$ denotes $(S_t, A_t, Z_{t+1}, Z_t),\ t=\{0, \cdots, T-1\}$. It can be observed that the discriminator $D_{\psi}$ is trained as a classifier to distinguish the expert demonstrations (labeled as 0) and generated samples (labeled as 1), and does not have a specially-designed structure like the discriminator $D_{\vartheta}$ in H-AIRL (Equation \eqref{equ:10}) so that it cannot recover the expert reward function. Note that, for H-GAIL, we only change the definition of $L^{IL}$ in the overall objective, i.e., Equation \eqref{equ:19}, with the one in Equation \eqref{equ:a35}, but keep the directed information objective $L^{DI}$ which is not included in the objective of Option-GAIL.